\documentclass{naturep}

\usepackage{amssymb}
\usepackage{amsmath}
\usepackage{graphicx}
\usepackage{algorithmicx}
\usepackage{algorithm}
\usepackage{subfigure}
\usepackage{subcaption}
\usepackage[noend]{algpseudocode}
\usepackage{bbm}
\usepackage{lineno}
\usepackage[margin=0.6in]{geometry}
\usepackage{ragged2e}
\usepackage{setspace}
\usepackage{longtable}
\usepackage{hyperref}
\usepackage{anyfontsize}
\usepackage{setspace}
\usepackage{float}
\usepackage{booktabs}
\usepackage{multirow}
\usepackage{xcolor}
\usepackage{blindtext}
\usepackage{tabularx}
\usepackage{caption}
\usepackage{array}
\usepackage{bbding}
\usepackage{pifont}
\usepackage{fontawesome}
\usepackage{xurl}
\usepackage{float}
\usepackage[section]{placeins}

\usepackage[normalem]{ulem}

\usepackage{subcaption}

\makeatletter
\let\saved@includegraphics\includegraphics
\AtBeginDocument{\let\includegraphics\saved@includegraphics}

\makeatother

\usepackage{enumitem, kantlipsum}

\title{Towards a general-purpose foundation model for fMRI analysis}

\begin{document}

\maketitle

\begin{spacing}{1.8}
\vspace{-10mm}
\noindent Cheng Wang$^{1}$, Yu Jiang$^{1}$, Zhihao Peng$^{1}$, Chenxin Li$^{1}$, Changbae Bang$^{2}$, Lin Zhao$^{3}$, Wanyi Fu$^{4}$,\\Jinglei Lv$^{5}$, Jorge Sepulcre$^{6}$, Carl Yang$^{7}$, Lifang He$^{8}$, Tianming Liu$^{3}$, Xue-Jun Kong$^{9}$,\\Quanzheng Li$^{10}$, Daniel S. Barron$^{11}$, Anqi Qiu$^{12}$, Randy Hirschtick$^{10}$, Byung-Hoon Kim$^{2}$, \\Hongbin Han$^{4*}$,Xiang Li$^{10,13*}$, Yixuan Yuan$^{1*}$\\
\end{spacing}
\vspace{-10mm}
\begin{spacing}{1.4}
\begin{affiliations}
 \item The Chinese University of Hong Kong, Hong Kong
 \item Yonsei University, Seoul, South Korea
 \item University of Georgia, Athens, GA
 \item Peking University Health Science Center, Beijing, China
 \item University of Sydney, Sydney, Australia
 \item Yale School of Medicine, New Haven, CT
 \item Emory University, Atlanta, GA
 \item Lehigh University, Bethlehem, PA
 \item Boston Children's Hospital, Boston, MA
 \item Massachusetts General Hospital, Boston, MA
 \item Brigham and Women's Hospital, Boston, MA
 \item Hong Kong Polytechnic University, Hong Kong
 \item Kempner Institute for Natural and Artificial Intelligence, Cambridge, MA
 
{*Corresponding authors}
\end{affiliations}
\end{spacing}

\footnotetext[1]{
Cheng Wang; 
Yu Jiang; 
Zhihao Peng; 
Chenxin Li; 
Changbae Bang; 
Lin Zhao; 
Wanyi Fu; 
Jinglei Lv; 
Jorge Sepulcre; 
Carl Yang; 
Lifang He; 
Tianming Liu; 
Xue-Jun Kong; 
Quanzheng Li; 
Daniel Barron; 
Anqi Qiu; 
Randy Hirschtick; 
Byung-Hoon Kim; 
Hongbin Han: \nolinkurl{hanhongbin@bjmu.edu.cn}; 
Xiang Li: \nolinkurl{xli60@mgh.harvard.edu}; 
Yixuan Yuan: \nolinkurl{yxyuan@ee.cuhk.edu.hk}.
}

\vspace{-5mm}
\begin{spacing}{1.0}

\section{Abstract} 
Functional MRI (fMRI) is crucial for studying brain function and diagnosing neurological disorders. However, existing analysis methods suffer from reproducibility and transferability challenges due to complex preprocessing pipelines and task-specific model designs. In this work, we introduce a \textbf{Neuro}imaging Foundation Model with \textbf{S}patial-\textbf{T}emporal \textbf{O}ptimized and \textbf{R}epresentation \textbf{M}odeling (NeuroSTORM) that learns generalizable representations directly from 4D fMRI volumes and enables efficient transfer to diverse downstream applications. Specifically, NeuroSTORM is pre-trained on 28.65 million fMRI frames from over 50,000 subjects, spanning multiple centers and ages 5 to 100. It combines an efficient spatiotemporal modeling design and lightweight task adaptation to enable scalable pre-training and fast transfer to downstream applications. Here we show that NeuroSTORM consistently outperforms existing methods across five downstream tasks, including demographic prediction, phenotype prediction, disease diagnosis, re-identification, and state classification. On two multi-hospital clinical cohorts with 17 diagnoses, NeuroSTORM achieves the best diagnosis performance while remaining predictive of psychological and cognitive phenotypes. These results suggest that NeuroSTORM could become a standardized foundation model for reproducible and transferable fMRI analysis.

\end{spacing}

\section{Introduction}
Functional Magnetic Resonance Imaging (fMRI) is a non-invasive imaging technique widely used to study brain function and structure~\cite{hearne2016functional}, and is also investigated in research contexts for potential use in assessment and monitoring of various neurological disorders~\cite{sorg2007selective,lewandowski2010polyamine}. Moreover, it provides robust technological support for brain signal decoding and human-computer interaction in brain-computer interface research~\cite{scotti2024reconstructing}. Driven by the advances in deep learning, recent works in fMRI analysis have achieved substantial improvements in the accuracy and efficiency of interpreting complex neural patterns from both resting-state fMRI (rsfMRI) and task fMRI (tfMRI) data ~\cite{he2022meta,kim2023large,kim2024learning,ortega2023brainlm,zhang2021identification,zhang2022predicting,zhang2024self,tian2020topographic}. 
Despite its theoretical and methodological progress, the field of fMRI remains fragmented across data formats, preprocessing pipelines, and analytic models, challenges recently summarized by~\cite{biswal2025history}. Specifically, analytic reproducibility is fundamental to neuroimaging research, and overlooking it can introduce bias into subsequent analyses~\cite{botviniknezer2023reproducibility}. While prior studies related to reproducibility and transferability have focused on identifying reliable biomarkers~\cite{cao2014test}, standardizing pipelines~\cite{waller2022enigma}, and validating test-retest reliability~\cite{noble2021guide}, most of them remain tailored to specific fMRI applications and lack the generalizability to other tasks. Thus, there is an opportunity to develop a model that is intrinsically generalizable across diverse experimental settings and conditions, which can then enable both reproducibility and transferability within a unified framework.

\begin{figure*}[!t]
\centering
\includegraphics[width=.98\linewidth]{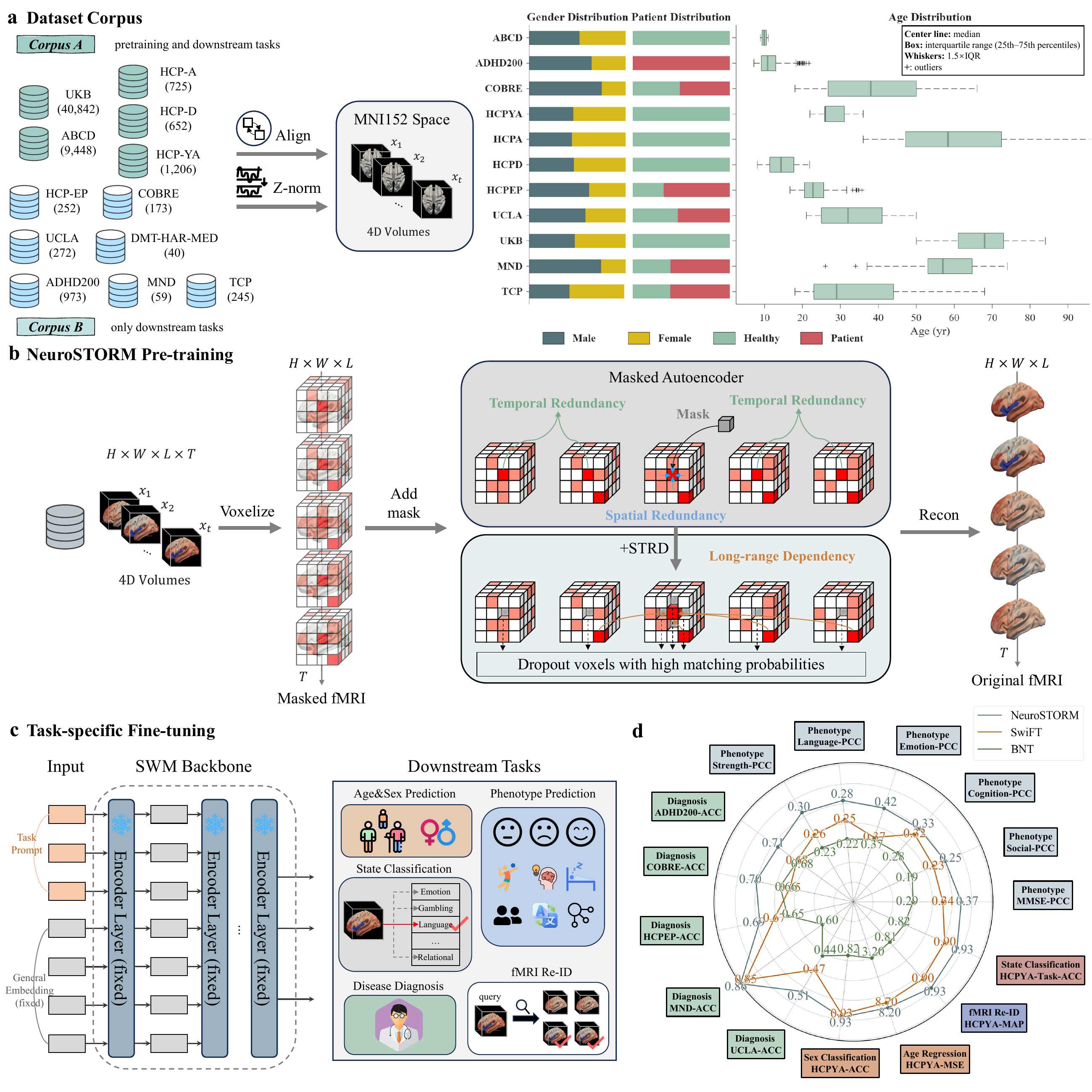}
\caption{Overview of our proposed NeuroSTORM framework. (a) Data corpus and preprocessing: The model is pre-trained on a collection of publicly available datasets, including over 50,000 rsfMRI and 16,000 tfMRI sequences. All data are aligned to 2mm MNI152 space to create standardized 4D volumes. Since we did not collect the complete metadata for the DMT-HAR-MED dataset, DMT-HAR-MED is not included in statistical information of the datasets. (b) NeuroSTORM pre-training: The model utilizes a masked autoencoder paradigm with a STRD module to enhance the learning of long-range spatiotemporal relationships. (c) Downstream tasks with fine-tuning: For evaluation, NeuroSTORM employs a SWM Backbone and TPT techique for efficient fine-tuning to various downstream tasks. The benchmark includes age and gender prediction, phenotype prediction, disease diagnosis, fMRI re-identification, and fMRI state classification. (d) Comprehensive performance evaluation: We systematically benchmark NeuroSTORM against previous ROI-based and Volume-based state-of-the-art models across a diverse set of downstream tasks. The radar chart illustrates NeuroSTORM's consistent performance improvements.}
\label{fig:main}
\end{figure*}

The emergence of foundation models~\cite{achiam2023gpt,touvron2023llama,xu2024whole,zhou2023foundation} presents a paradigm-shifting framework by scalable learning across tasks and improved robustness through large-scale pre-training and adaptable architectures. Foundation models were initially developed for natural language processing tasks~\cite{achiam2023gpt,touvron2023llama}, where models such as ChatGPT demonstrated remarkable multitask capabilities by training on web-scale text corpora. This success has inspired analogous developments in the medical domain~\cite{zhang2024challenges}, where foundation models are being applied to address challenges such as anatomical variability and limited annotated data. For example, RETFound~\cite{zhou2023foundation} establishes a foundation for retinal imaging using self-supervised learning on 1.6 million unlabeled images. Similarly, AnyStar~\cite{dey2024anystar} demonstrates that domain-randomized generative models can achieve cross-modal 3D instance segmentation of star-convex anatomical structures, such as the tumor boundary in MRI scans, without relying on annotated training data. Foundation models enhance transferability by balancing generalizable feature extraction with domain-specific pattern recognition, adapting to the unique requirements of each imaging modality. They also improve reproducibility by capturing noise-resilient patterns, often through self-supervised reconstruction or contrastive-based pre-training frameworks. These approaches can reduce sensitivity to acquisition variations and mitigate the variability introduced by preprocessing pipelines~\cite{waller2022enigma}, while preserving meaningful neurobiological information.

However, different from other data modalities, developing foundation models for fMRI data remains fundamentally challenging. Firstly, the raw 4D fMRI signal, comprising up to $10^{6}$ voxels per scan, poses severe computational and optimization bottlenecks. Prior approaches typically reduce dimensionality by projecting data onto pre-defined brain atlases or connectomes~\cite{he2022meta,kim2023large,kim2024learning,ortega2023brainlm,zhang2021identification,zhang2022predicting,zhang2024self,tian2020topographic}. However, these operations result in irreversible information loss and impose structural biases that hinder generalizability across populations~\cite{kim2023swift,malkiel2022self}. Secondly, due to the high spatiotemporal redundancy in 4D fMRI volumes, we observe that standard Masked Autoencoders (MAEs) which are widely used for foundation model design struggle to learn informative representations, as masked voxels can often be trivially reconstructed from their spatial or temporal neighbors.

In line with recent calls for scalable, generalizable frameworks for fMRI~\cite{biswal2025history}, we introduce the \textbf{Neuro}imaging Foundation Model with \textbf{S}patial-\textbf{T}emporal \textbf{O}ptimized and \textbf{R}presentation \textbf{M}odeling (NeuroSTORM), a general-purpose fMRI foundation model designed to enhance reproducibility and transferability through large-scale pre-training and architectural innovation (Fig.~\ref{fig:main}(a)). 
To enhance computational efficiency in 4D fMRI processing, we introduce a Shifted-Window Mamba (SWM) backbone, which combines linear-time state-space modeling with shifted-window mechanisms to reduce complexity and GPU memory usage. During pre-training, we propose a Spatiotemporal Redundancy Dropout (STRD) module (Fig.~\ref{fig:main}(b)) for effective learning of inherent characteristics in fMRI data, thereby improving the model's robustness and reproducibility. For downstream task adaptation, our Task-specific Prompt Tuning (TPT) strategy (Fig.~\ref{fig:main}(c)) employs a minimal number of trainable, task-specific parameters when fine-tuning NeuroSTORM for new tasks. This provides a simple and integrated approach to applying NeuroSTORM across diverse applications. 
The corpus used to pre-train NeuroSTORM integrates three large-scale neuroimaging datasets: UK Biobank~\cite{palmer2007uk} (40,842 participants), ABCD~\cite{casey2018adolescent} (9,449 children), and the HCP datasets~\cite{van2013wu} (HCP-YA, HCP-A, and HCP-D; totaling over 2,300 subjects). Spanning diverse demographics (ages 9-100), clinical conditions, and acquisition protocols, the corpus ensures broad biological and technical variation. 

To validate the performance and transferability of NeuroSTORM, we established a comprehensive fMRI analysis benchmark, including five downstream tasks: age and (reported) gender prediction, phenotype prediction, disease diagnosis, fMRI re-identification, and fMRI state classification. Notably, we assessed the clinical applicability of NeuroSTORM on two clinical datasets from hospitals in both the United States and Australia, Transdiagnostic Connectome Project (TCP)~\cite{chopra2024transdiagnostic} and Motor Neuron Disease (MND)~\cite{chang2025fmri}. The TCP dataset consists of 245 participants from Brain Imaging Center of Yale University or McLean Hospital in the United States, including both healthy controls and individuals covering a range of psychiatric disorders. The MND dataset includes 36 participants diagnosed with Amyotrophic Lateral Sclerosis and 23 controls, collected at the Royal Brisbane and Women's Hospital in Australia. NeuroSTORM outperforms or matches state-of-the-art models across all five tasks, demonstrating strong transferability in diverse applications. Moreover, we simulate a data-scarce scenario by limiting the proportion of fine-tuning data in order to evaluate label efficiency. NeuroSTORM demonstrates minor performance degradation on most datasets. Notably, we evaluated NeuroSTORM's ability to predict clinical phenotypes and perform disease classification in clinical datasets, with experimental results highlighting its clinical value and transferability.

Overall, we present a foundation model for fMRI analysis that achieves outstanding performance across five downstream tasks, enabling for large-scale fMRI studies with enhanced reproducibility and transferability. We have open-sourced a GitHub repository (github.com/CUHK-AIM-Group/NeuroSTORM), which serves as a general-purpose fMRI analysis platform. This repository includes tools for fMRI preprocessing, trainers for both pre-training and fine-tuning, a benchmark suite for diverse fMRI tasks, implementations of NeuroSTORM, as well as other commonly used fMRI analysis models. Detailed guidelines are provided for adding custom preprocessing procedures, pre-training methods, fine-tuning strategies, new downstream tasks, and additional models to the platform. This platform enhances NeuroSTORM's reproducibility and transferability while advancing fMRI research.

\section{Results}
The downstream task of age and (reported) gender prediction utilizes HCP-YA~\cite{van2013wu}, HCP-A, HCP-D, UKB~\cite{palmer2007uk} and ABCD~\cite{casey2018adolescent} datasets. Phenotype prediction is based on HCP-YA and TCP~\cite{chopra2024transdiagnostic} dataset. Disease diagnosis draws on multiple datasets, including HCP-EP~\cite{van2013wu}, ADHD200~\cite{brown2012adhd}, COBRE~\cite{calhoun2012exploring}, UCLA~\cite{poldrack2016phenome} and MND~\cite{chang2025fmri}. The fMRI re-identification task is evaluated on the HCP-YA. fMRI state classification task uses the HCP-YA and DMT-HAR-MED~\cite{meling2024meditating} dataset. No annotations are included in the NeuroSTORM pre-training corpus. We compare NeuroSTORM against with baselines from two major paradigms: (i) \emph{ROI-based methods} that convert the 4D fMRI into 2D region-of-interest (ROI) connectomes, includes BrainGNN~\cite{li2021braingnn}, BrainNet Transformer (BNT)~\cite{kan2022brain}, LG-GNN~\cite{zhang2022classification}, Com-brainTF~\cite{bannadabhavi2023community}, IBGNN~\cite{cui2022interpretable}, Brain-JEPA ~\cite{dong2024brain} and BrainLM~\cite{ortega2023brainlm}; and (ii) \emph{volume-based methods} that directly precess voxel-wise 4D volumes, TFF~\cite{malkiel2022self} and SwiFT~\cite{kim2023swift}. SwiFT is particularly relevant to our work as it also employs an end-to-end backbone tailored for spatio-temporal 4D inputs. All baselines are re-implemented or run with authors' code under identical preprocessing to ensure a fair comparison. In all experiments, the downstream task datasets are split into training, validation, and testing sets in an 8:1:1 ratio. The same dataset split and training details are applied to all comparison methods. Additionally, we evaluate NeuroSTORM's label efficiency in data-scarce scenarios by employing varying percentages of fine-tuning data in the datasets.

\begin{figure*}[!t]
\centering
\includegraphics[width=\textwidth]{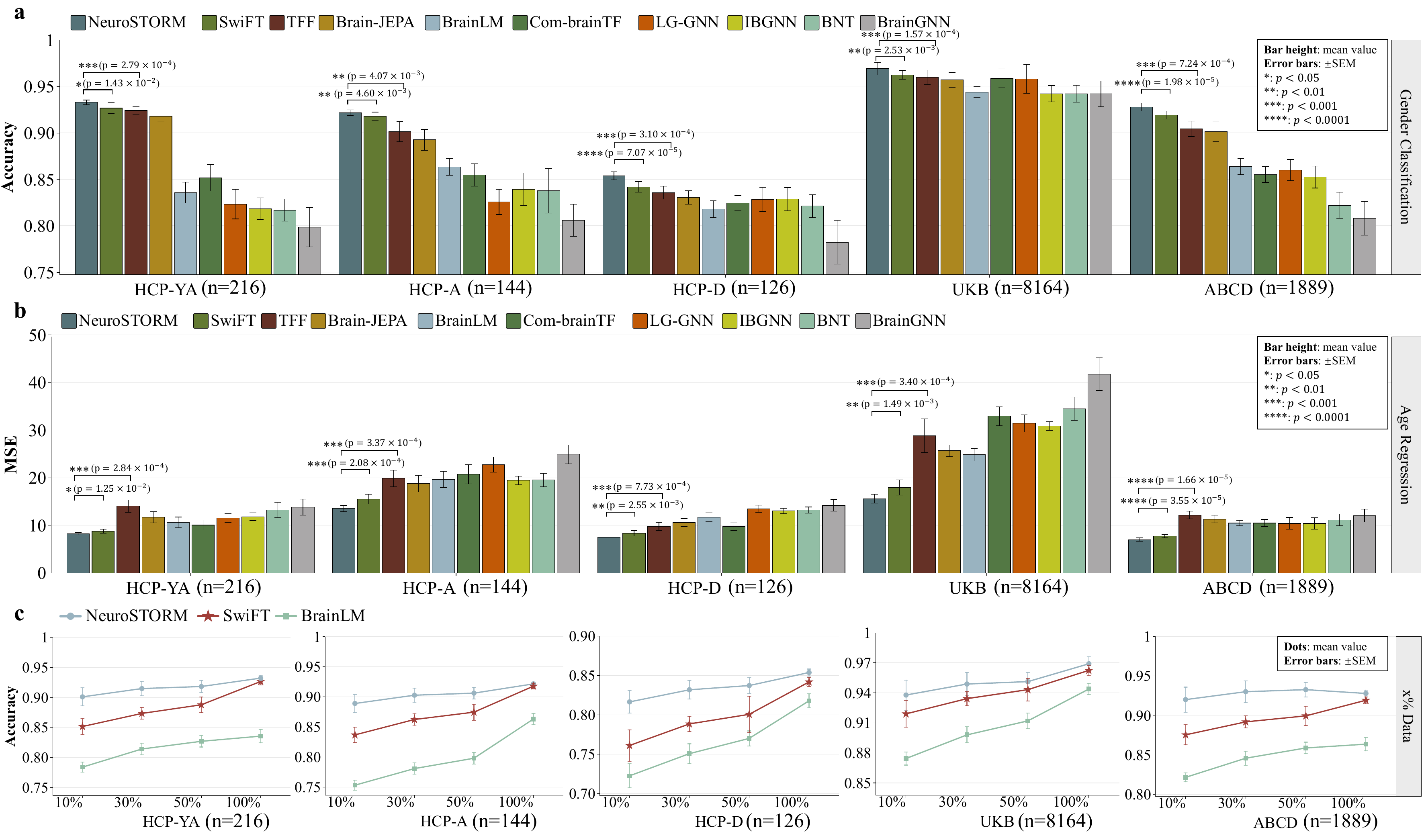}
\caption{Evaluation of NeuroSTORM's performance in (reported) gender classification and age regression tasks. (a) Gender classification performance: NeuroSTORM consistently outperformed competing ROI-based and volume-based methods across datasets, including HCP-YA, HCP-A, HCP-D, UKB and ABCD. (b) Age regression error: In age regression tasks, NeuroSTORM achieved the lowest error rates, surpassing competing methods across multiple datasets. (c) Label efficiency results: NeuroSTORM demonstrated strong adaptability when trained with limited data. Remarkably, even with only 10\%-50\% of the training data, it maintained competitive performance in both age and gender prediction tasks. With increasing training data proportions, its performance steadily improved, achieving optimal results with full datasets. Variances were estimated from five technical replicates.Pairwise significance markers were computed using a two-sided paired t-test (\texttt{ttest\_rel}) without multiple-comparison correction, and the corresponding P-value is annotated.}
\label{fig:task1}
\end{figure*}

\subsection{Age and Gender Prediction}\label{sec2.1}
To evaluate the transferability of NeuroSTORM, we first assessed its performance in predicting age and gender from rsfMRI sequences compared to state-of-the-art fMRI analysis methods. Age and gender are fundamental sociodemographic variables correlating with structural brain changes, making this a fundamental fMRI analysis task. Since the dataset metadata includes only reported gender, the prediction target in this downstream task is that label. NeuroSTORM consistently outperformed both ROI-based methods and volume-based approaches across all benchmark datasets. On HCP-YA~\cite{van2013wu} it reached a gender-classification accuracy of 93.28\%, surpassing the best volume-based method SwiFT~\cite{kim2023swift} at 92.66\% and the best ROI-based method Brain-JEPA at 91.80\%. Similar superiority is observed in HCP-A (92.16\% vs 91.78\% for SwiFT and 89.24\% for Brain-JEPA) and HCP-D (85.38\% vs 84.18\% for SwiFT and 83.04\% for Brain-JEPA). For age prediction on ABCD, NeuroSTORM reduced Mean Square Error (MSE) to 7.0 compared to SwiFT~\cite{kim2023swift}'s 7.72 (best volume baseline) and LG-GNN's 10.40 (best ROI baseline). While traditional ROI-based methods showed advantages in small datasets due to lower dimensionality and higher signal-to-noise ratios, NeuroSTORM's architecture ultimately surpassed them through effective integration of structural and functional features (Fig.~\ref{fig:task1}).

NeuroSTORM also exhibited strong label efficiency in data-scarce scenarios. For instance, on HCP-D, training with only 50\% of the available labels still produced 83.7\% accuracy, clearly ahead of SwiFT's 80.1\% and BrainLM's 77.0\%. When the training set was limited to 30\%, NeuroSTORM retained 83.2\% accuracy, losing less than 3\% absolute performance relative to full-data training, whereas SwiFT and BrainLM dropped by more than 5\% and 6\%, respectively. These results show that the pre-training strategy of NeuroSTORM enables efficient knowledge transfer and robust generalization even when labeled data are scarce.

\subsection{Phenotype Prediction}\label{sec2.2}
To evaluate the effectiveness of NeuroSTORM in capturing functional brain mapping and connectivity, we conducted phenotype prediction experiments using the HCP-YA~\cite{van2013wu} and TCP~\cite{chopra2024transdiagnostic} datasets. HCP-YA dataset contains normalized scores across mental (P01), social (P02), cognitive (P03), emotional (P04), language (P05), and physical (P06) domains. The TCP dataset incorporates a comprehensive suite of clinical phenotypes (P07-P16), including measures of anxiety, depression, stress, cognitive performance, and personality traits, specifically designed for individuals with a diverse range of psychiatric disorders. The specific phenotype names used can be found in the appendix.

\begin{figure*}[!t]
\centering
\includegraphics[width=\textwidth]{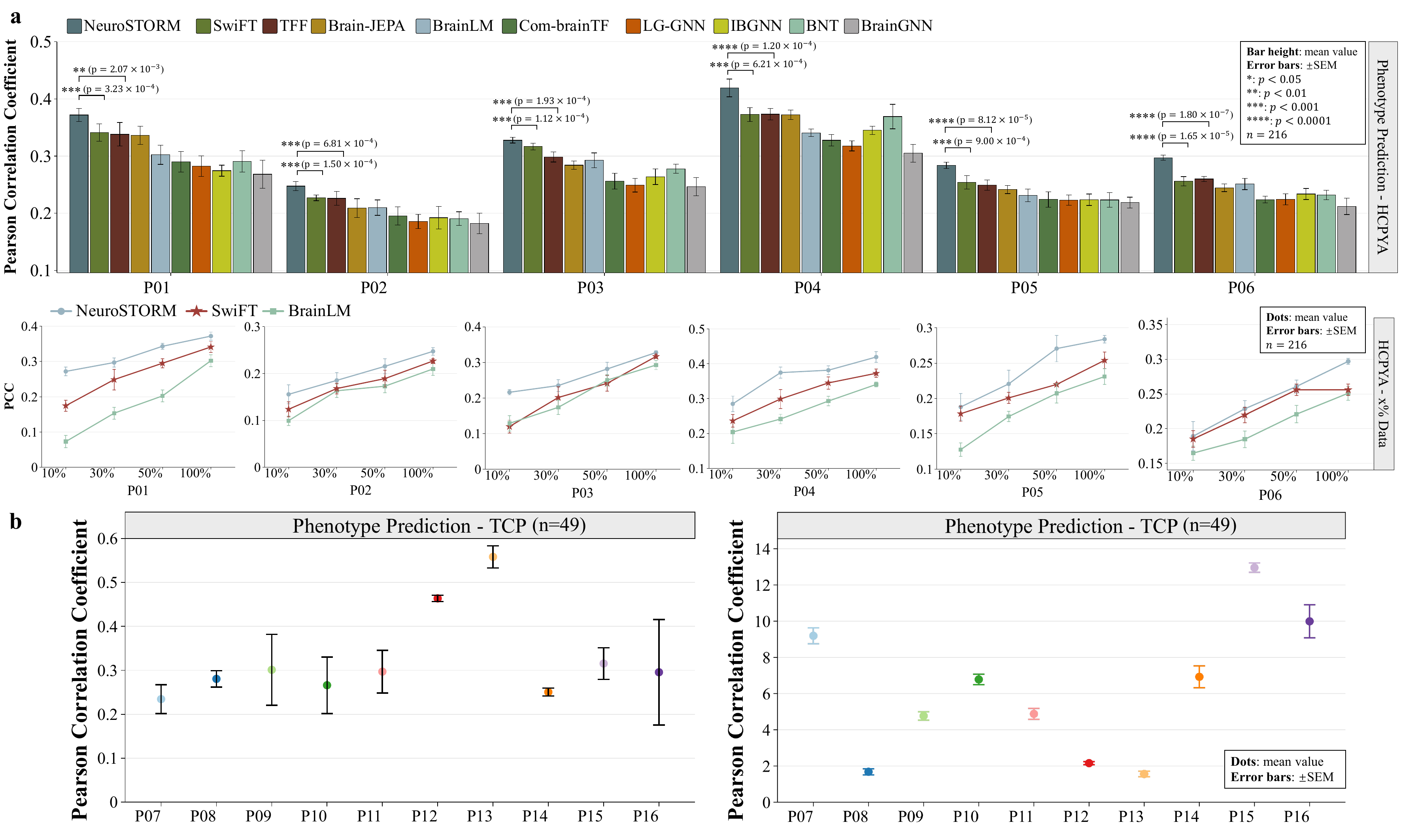}
\caption{Performance evaluation of NeuroSTORM in phenotype prediction task. (a) HCP-YA dataset: NeuroSTORM demonstrates superior Pearson Correlation Coefficients (PCC) across diverse phenotype scores, including MMSE Score (P01), Social Task Performance (P02), Cognitive Total Score (Age Adjusted) (P03), Emotion Task Accuracy (P04), Language Task Accuracy (P05), and Strength Score (Age Adjusted) (P06). Even in data-scarce scenarios (10\%-50\%), NeuroSTORM maintains competitive PCC performance. (b) TCP dataset: NeuroSTORM is evaluated for its ability to predict crucial disease-related scores in subjects with psychiatric disorders, including Anxiety Sensitivity (P07), CGI Severity Score (P08), DASS Anxiety Score (P09), DASS Stress Score (P10), PANSS General Score (P11), PANSS Negative Symptoms (P12), PANSS Positive Symptoms (P13), NEO Agreeableness Score (P14), TCI Harm Avoidance Score (P15), and TCI Cautiousness Score (P16). Variances were estimated from five technical replicates. Pairwise significance markers were computed using a two-sided paired t-test (\texttt{ttest\_rel}) without multiple-comparison correction, and the corresponding P-value is annotated.}
\label{fig:task2}
\end{figure*}

As shown in Fig.~\ref{fig:task2}, NeuroSTORM demonstrated improved performance compared to ROI-based and volume-based methods across various scores. For MMSE (P01) it reaches 0.372, which is 9.1\% higher than the strongest volume baseline SwiFT (0.341) and 31.5\% higher than the best ROI baseline Brain-JEPA (0.336). Consistent gains are observed for Social Task Performance (P02, 0.247 vs 0.227 for SwiFT), Cognitive Total Score (P03, 0.328 vs 0.317 for SwiFT), Emotion Task Accuracy (P04, 0.419 vs 0.373 for SwiFT), Language Task Accuracy (P05, 0.284 vs 0.254 for SwiFT) and Strength Score (P06, 0.297 vs 0.260 for TFF). Label-efficiency curves reveal a slower degradation for NeuroSTORM: on P01 it loses only 7.8\% when the label budget drops from 100\% to 50\%, whereas SwiFT and BrainLM drop 13.5\% and 33.1\%, respectively. NeuroSTORM reaches 0.348 PCC, surpassing SwiFT trained with the full dataset (0.341). For P04, NeuroSTORM attains 0.375 with 30\% labels, which is close to SwiFT's 0.373 with 100\%, and rises to 0.381 with 50\% labels, illustrating a consistently flatter performance curve than competing methods.

For the TCP dataset, NeuroSTORM was evaluated on its ability to predict clinical phenotypes associated with psychiatric disorders. The model's performance varied across different phenotypes, achieving PCC ranging from 0.234 to 0.558. Specifically, highest correlations are obtained for PANSS Positive Symptoms (P13, PCC=0.558) and PANSS Negative Symptoms (P12, PCC=0.464). Lower correlations are observed for Anxiety Sensitivity (P07, PCC=0.234) and TCI Cautiousness (P16, PCC=0.296). Mean-absolute error ranges from 1.455 (P13) to 13.23 (P15), reflecting the varying difficulty of the prediction tasks. These results confirm that NeuroSTORM can model a broad spectrum of clinically relevant traits, though performance naturally varies with the specific target.

\begin{figure*}[!t]
\centering
\includegraphics[width=.99\linewidth]{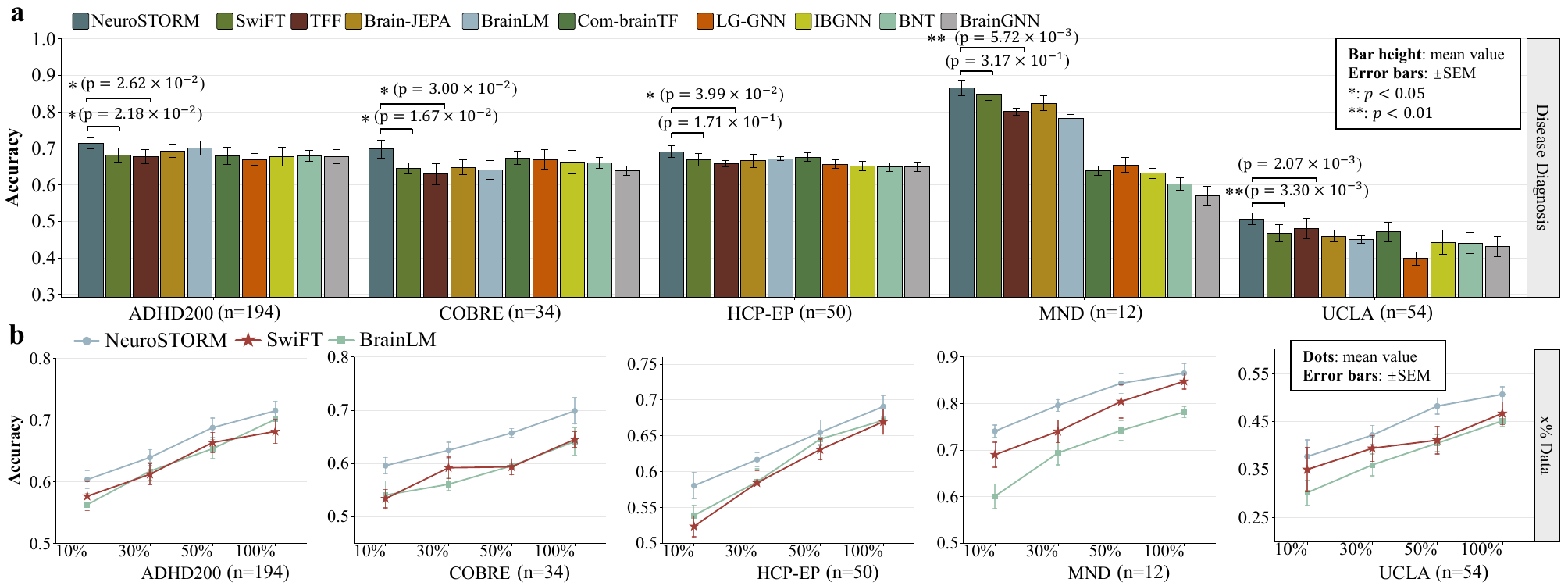}
\caption{Evaluation of NeuroSTORM's performance on disease diagnosis task. (a) Classification accuracy on ADHD200, COBRE, HCP-EP, MND and UCLA shows that NeuroSTORM consistently outperforms all ROI-based and volume-based baselines, highlighting its strong generalisability across neurological and psychiatric disorders. (b) Label efficiency analysis demonstrates that NeuroSTORM maintains robust performance even when trained with only a fraction of the fine-tuning data, underscoring its suitability for data-scarce scenarios. Variances were estimated from five technical replicates. Pairwise significance markers were computed using a two-sided paired t-test (\texttt{ttest\_rel}) without multiple-comparison correction, and the corresponding P-value is annotated.}
\label{fig:task3}
\end{figure*}

\subsection{Disease Diagnosis}\label{sec2.3}
We evaluated NeuroSTORM's performance against state-of-the-art methods on the Disease Diagnosis task across multiple datasets, representing different neurological and psychiatric conditions. These datasets include HCP-EP~\cite{van2013wu}, ADHD200~\cite{brown2012adhd}, COBRE~\cite{calhoun2012exploring}, UCLA~\cite{poldrack2016phenome} and motor neuron disease (MND)~\cite{chang2025fmri}. As shown in Fig.~\ref{fig:task3}a, NeuroSTORM consistently outperforms all baseline methods across datasets. On ADHD200 (healthy vs ADHD) it achieves 71.46\%, 1.38\% higher than the best competitor BrainLM. For the COBRE benchmark NeuroSTORM reaches 69.84\%, surpassing Com-brainTF by 2.44\%. In early-psychosis detection (HCP-EP) the model obtains 69.06\%, a 1.44\% gain over Com-brainTF. On the MND cohort NeuroSTORM scores 86.50\%, improving on SwiFT by 1.74\%. The largest margin was observed on the UCLA dataset, where NeuroSTORM attained 50.66\%, 2.62\% above TFF and more than 4\% ahead of all volume baselines. These results demonstrate NeuroSTORM's robust and generalizable diagnostic capability across diverse brain disorders~\cite{jablensky2016psychiatric,joyce2023explainable}.

For label efficiency, we systematically evaluated NeuroSTORM's sample efficiency using progressively reduced training subsets (Fig.~\ref{fig:task3}b). On ADHD200 the model retains 68.7\% accuracy with only 50\% of the labels, already surpassing Com-brainTF trained on all data (68.0\%) and nearly matching the best volume baseline BrainLM (70.1\%). On the COBRE dataset NeuroSTORM delivers 62.4\% accuracy with 30\% of the labels, only 2.1\% below SwiFT's full-data score of 64.5\% and close to BrainLM's 64.1\%. These examples illustrate that NeuroSTORM's performance degrades more gradually as labelled data decrease, and that it maintains comparable accuracy to state-of-the-art methods in data-scarce scenarios.

\subsection{fMRI Re-identification}\label{sec2.4}
To evaluate the discriminative strength of NeuroSTORM's embedding space, we follow a closed-set fMRI re-identification protocol in which the held-out data are split into a query set and a non-overlapping gallery set, with the gallery size fixed at 100. During inference, each 4D sequence is forwarded once to obtain a single $\ell_{2}$-normalized feature vector, and an exhaustive nearest-neighbor search is performed in the gallery. Retrieval quality is reported with the standard Rank-1 Accuracy and mean Average Precision (mAP) over the entire query set. This experimental design offers a rigorous test of how well the foundation model retains subject-specific cues~\cite{finn2015functional} and supports large-scale similarity search in high-dimensional fMRI data.

\begin{figure*}[!t]
\centering
\includegraphics[width=\textwidth]{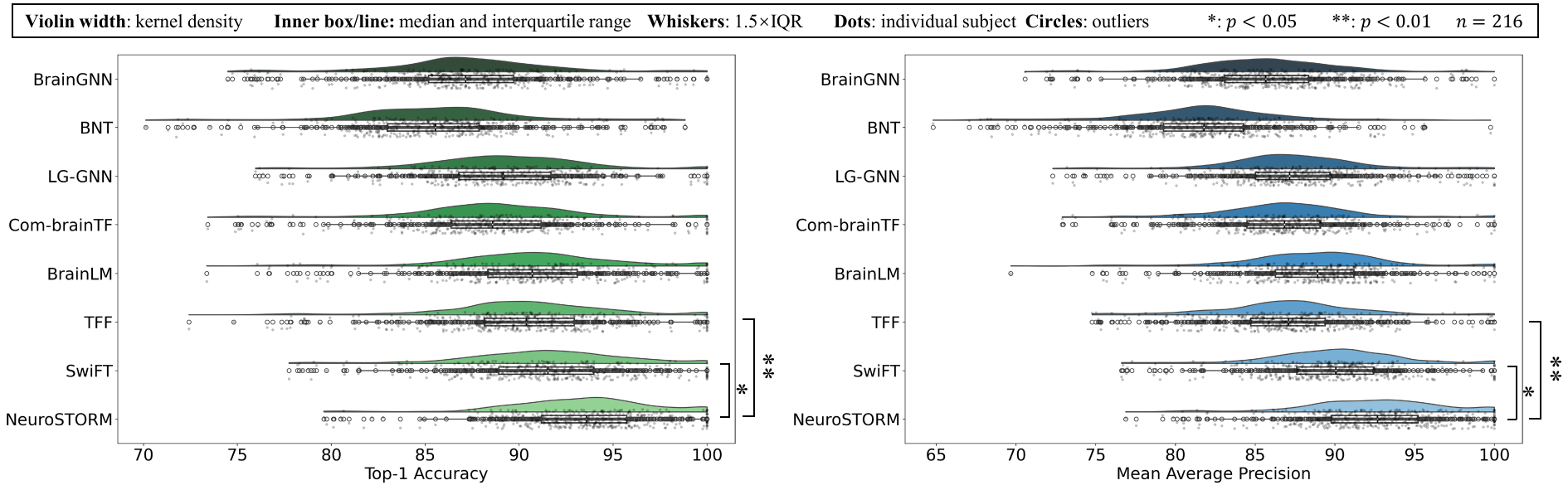}
\caption{Evaluation of fMRI re-identification on HCP-YA with the gallery size fixed at 100. Under a closed set protocol, each 4D sequence is embedded once into an $\ell_{2}$-normalized feature vector and matched via exhaustive nearest neighbor search in the gallery, with retrieval quality reported by Rank-1 Accuracy and mAP.}
\label{fig:task4}
\end{figure*}

As shown in Fig. \ref{fig:task4}, NeuroSTORM achieves 93.1\% Rank-1 accuracy and 92.4\% mAP under the same closed-set protocol, surpassing all competing methods. Compared with the strongest baseline SwiFT, the gains are +1.5\% in Rank-1 and +3.0\% in mAP. ROI-based baselines show lower performance. For example, the best of ROI-based model, BrainLM, reaches 90.8\% Rank-1 and 88.2\% mAP, whereas Com-brainTF attains 89.0\% and 87.4\%. The other methods obtain even lower scores. These margins suggest that the compact $\ell_{2}$-normalized embeddings produced by our shifted scanning strategy on a Mamba backbone preserve strong subject-specific signatures. The spatial-temporal optimized pre-training coupled with task-specific prompt tuning further supports reliable one-to-many retrieval in high-dimensional fMRI space.

\subsection{fMRI State Classification}\label{sec2.5}
In this study, we evaluated the state classification performance of NeuroSTORM in the HCP-YA~\cite{van2013wu} and DMT-HAR-MED~\cite{meling2024meditating} dataset. The primary objective of the task is to accurately identify the cognitive state or experimental condition. The HCP-YA dataset comprises seven distinct functional tasks: Emotion, Gambling, Language, Motor, Relational, Social, and Working Memory (WM). Each task is designed to activate different brain networks associated with unique cognitive states. The DMT-HAR-MED dataset includes resting-state fMRI from 40 meditators collected before and after a retreat, with participants receiving either DMT-harmine or placebo.

\begin{figure*}[!t]
\centering
\includegraphics[width=\textwidth]{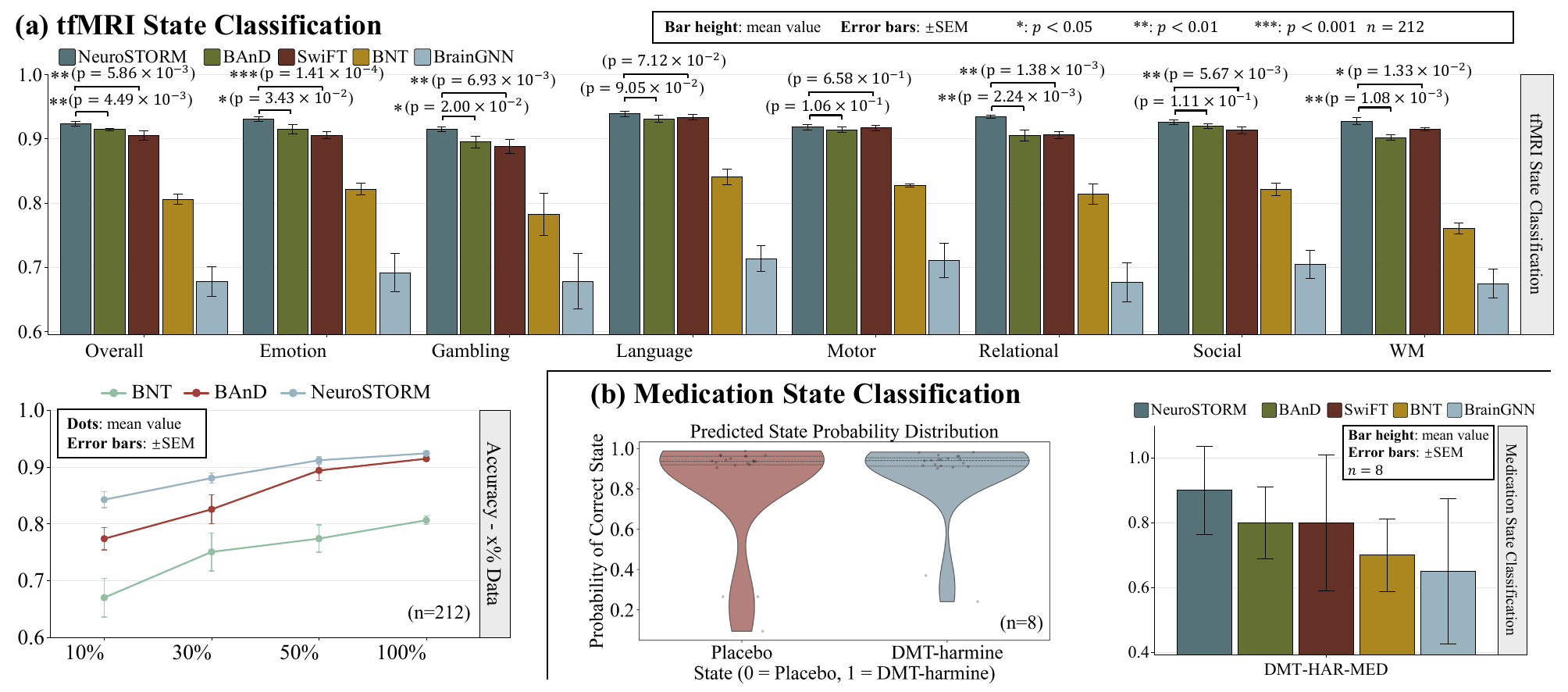}
\caption{Evaluation of Performance for fMRI state classification task. (a) tfMRI state classification results: NeuroSTORM demonstrated superior accuracy across all tasks compared to other ROI-based and volume-based methods, achieving an overall accuracy of 92.64\%. (b) Medication state classification results: NeuroSTORM distinguished DMT-harmine from placebo with 87.5\% overall accuracy. The violin plots display the distribution of predicted probabilities assigned to the true class for each subject; correct predictions cluster at high-confidence values while errors occur at markedly lower probabilities, indicating well-calibrated predictions. Variances were estimated from five technical replicates. Pairwise significance markers were computed using a two-sided paired t-test (\texttt{ttest\_rel}) without multiple-comparison correction, and the corresponding P-value is annotated.}
\label{fig:task5}
\end{figure*}

NeuroSTORM exhibited superior performance across all task states (Fig.~\ref{fig:task5}(a)), highlighting its ability to effectively generalize features. Our framework achieved an overall accuracy of 92.40\%, outperforming the strongest volume-based baseline BAnD (91.46\%) and showing clear margins over the best ROI baselines BNT (80.62\%). Task-specific comparisons further evaluated the model capacity in complex cognition tasks. NeuroSTORM achieved 93.90\% accuracy in Language (+0.78\% over BAnD) and 93.06\% on Emotion (+1.56\%). Additionally, we investigated NeuroSTORM's performance robustness in data-scared scenarios. NeuroSTORM retained 91.2\% overall accuracy with 50\% of the labels, matching BAnD's full-data score of 91.5\% and remaining far ahead of the strongest ROI baseline BNT at full data (80.6\%). Even with only 10\% of the labels NeuroSTORM still reached 84.2\%, whereas BAnD fell to 77.4\% and BNT to 67.0\%. These results underscore the label efficiency of NeuroSTORM in data-scared settings.

In the DMT-HAR-MED~\cite{meling2024meditating} dataset, we evaluated NeuroSTORM for medication state classification between DMT harmine and placebo. As shown in Fig.~\ref{fig:task5}(b), NeuroSTORM achieved 90.0\% average accuracy in five independent experiments, demonstrating effective detection of medication related functional changes in resting state. Confidence analysis shows that correct predictions have probability outputs concentrated at high values, whereas incorrect predictions have markedly lower confidence. These results confirm that NeuroSTORM delivers high accuracy together with high confidence and stable predictive behavior in medication state classification, highlighting robustness and practical utility in resting state drug intervention settings.

\section{Discussion}\label{sec3}
Our study addresses two fundamental challenges in fMRI research, the limited model reproducibility and transferability, through systematic innovations in data curation, architecture design, and task benchmarking. The construction of the large multi-source fMRI pre-training corpus ($>50,000$ scans) combined with the STRD module and SWM architecture establishes NeuroSTORM as a versatile foundation model. NeuroSTORM directly processes raw 4D volumes while mitigating the influence of scanner artifacts and physiological noise. By formulating fMRI feature learning as spatiotemporal reconstruction invariant to population-level heterogeneity, we demonstrate that large-scale pre-training can simultaneously mitigate fundamental issues in test-retest reliability and cross-task generalization. The comprehensive benchmark spanning five domains (sociodemographic prediction, phenotype estimation, disease diagnostics, fMRI re-identification, and state classification) systematically evaluates general-purpose fMRI modeling capabilities, covering clinical, neuroscientific, and BCI applications.

Unlike ROI-based approaches that discard anatomical context through atlas parcellation~\cite{he2022meta,kim2023large}, NeuroSTORM operates directly on whole-brain 4D fMRI volumes, thereby preserving the fine-grained spatiotemporal relationships that are critical for modelling neural activity. This volume-level strategy need substantial computational efficiency, so we adopt a Shifted-Window Mamba backbone that can process high-dimensional inputs within practical GPU memory limits. During downstream adaptation we further reduce the computational footprint through task-prompted tuning, which updates fewer than 5\% of the parameters while maintaining full model capacity. A second technical challenge is the intrinsic redundancy of fMRI signals. The STRD module addresses this issue by selectively masking spatially and temporally redundant voxels during masked pre-training and then reconstructing both informative brain signals and subject-specific physiological noise. These combined innovations enable NeuroSTORM to learn long-range spatiotemporal dependencies and to produce a transferable feature space that generalises across tasks, datasets and computational budgets. Consequently, NeuroSTORM provides a flexible foundation for future fMRI research and downstream modelling efforts under diverse real-world conditions.

Moreover, we design a fMRI analysis benchmark which meets key requirements for evaluating foundation models in neuroimaging: (1) task diversity (regression/classification across neurodevelopmental, psychiatric, and neurodegenerative conditions), (2) feature-space validity (demonstrated by the structured embeddings uncovered in the fMRI re-ID task), and (3) clinical transferability (efficacy in low-data regimes through prompt tuning). The inclusion of re-identification and state classification tasks further tests the model's capacity to decode fine-grained semantic content, which is absent in the benchmark of previous fMRI analytical models.

NeuroSTORM demonstrates statistically superior performance across all benchmarking tasks compared to state-of-the-art methods. For disease diagnosis, it achieves 69.06\% accuracy on schizophrenia detection (HCP-EP) and 50.66\% accuracy differentiating ADHD subtypes (UCLA), outperforming ROI-based approaches by 1.44\% and 3.54\% respectively. The model attains 93.28\% gender classification accuracy (HCP-YA) with 8.24 MAE in age prediction, reducing errors by 5.5\% versus SwiFT~\cite{kim2023swift}. Phenotype prediction reveals strong alignment with established biomarkers, such as the 0.419 PCC for emotion task in HCP-YA dataset. fMRI state classification reaches 92.4\% accuracy across seven different states, confirming effective adaptation to both resting-state and task activity patterns. Notably, we observe that ROI-based methods exhibit lower performance compared to volume-based approaches, due to the loss of state-related information during downsampling operations; this finding is also consistent with reported results in previous works~\cite{nguyen2020attend}. Furthermore, NeuroSTORM surpasses the transformer-based SwiFT on the vast majority of tasks, this comparison validates the effectiveness of our SWM backbone; Sec.~\ref{sec:methods} further shows that NeuroSTORM attains these gains while consuming less GPU memory, underscoring the computational efficiency of the proposed architecture. This improvement benefits from the window mixing and state space model design that captures long-range spatiotemporal dependencies without incurring the quadratic complexity of full self-attention.

NeuroSTORM's parameter-efficient prompt tuning requires $\leq5\%$ tuned weights while maintaining efficient data utilization. For instance, NeuroSTORM achieves $91.18\%$ state classification accuracy with $50\%$ of the training data in HCP-YA dataset, which is a critical advantage in data-scarce scenarios. When the available data are further reduced to $30\%$, the model still attains $88.04\%$ overall accuracy, only $4.36$ percentage points below the $92.40\%$ obtained with the full dataset. Even with merely $10\%$ of the data, NeuroSTORM retains $84.24\%$ overall accuracy, and every individual task remains above $83\%$. These findings confirm that the proposed prompt-tuning scheme preserves balanced generalization across cognitive states and degrades gracefully under severe data-limitation settings.

\textbf{Limitations and Future Directions}. Three key limitations of NeuroSTORM shall guide future research directions. First, while rsfMRI data enables broad applicability, tfMRI integration (less than 20\% of pre-training corpus) could enhance decoding specificity for cognitive operations. Second, the current architecture learns spatial correlation patterns through voxel masking and reconstruction without any anatomical priors, potentially losing local connectivity knowledge discovered in previous works. Incorporating graph neural network using individual-specific brain connectivity network may enhance neurobiological interpretability. Finally, although demonstrating cross-scanner generalization, population bias in training data (predominantly North American/European cohorts) may constrain global applicability.

Our fMRI foundation model opens up revolutionary research directions by integrating neuroscientific discoveries. Through self-supervised learning on large fMRI datasets, the model develops noise-robust feature representation that has generalization ability to detect clinical brain state fluctuations for patients with different health conditions. For cognitive neuroscience applications, NeuroSTORM has the potential to offer a system-level analytical platform that enables researchers to explore whole-brain activity through task-independent features, which could help reveal previously unrecognized functional circuits and their cognitive associations. Additionally, the framework natively accommodates multimodal extensions, such as DTI structural connectivity and EEG signals, potentially advancing toward comprehensive brain modeling.

\section{Methods} \label{sec:methods}
\textbf{Shifted Window Mamba Backbone}. NeuroSTORM is built on a SWM Backbone network, which efficiently processes raw voxel-wise fMRI volumes. The Shifted Windowing mechanism ~\cite{cao2022swin} optimizes the computation of self-attention by dividing the input into manageable windows, reducing complexity from quadratic to linear. This is achieved by alternating between regular and shifted window partitions, allowing for cross-window connections without sacrificing computational efficiency. This method enhances the model's ability to capture spatial relationships across different regions of the brain. We utilize a variant of the State Space Model (SSM)~\cite{gu2021efficiently,gu2021combining} known as the Mamba model ~\cite{gu2023mamba}, which offers improved context selection and the ability to compress historical information. By integrating these two powerful mechanisms, SWM Backbone provides a robust framework for analyzing high-dimensional fMRI data, capturing both local and global patterns effectively.

Specifically, the input sequence is first converted to sequence embeddings by a patch embedding layer, which are then transformed into the hidden space. In the encoder, multiple SSM blocks and patch merging layers extract hierarchical feature representations with four stages at different scales: \(\frac{1}{4}, \frac{1}{8}, \frac{1}{16}, \frac{1}{32}\), where the number of blocks in each stage are \((L_1, L_2, L_3, L_4)\), and the hidden dimensions are \((C_1, C_2, C_3, C_4)\), respectively. Patch merging layers reduce the number of tokens and increase the feature dimension. We follow the Mamba model~\cite{gu2023mamba} to offer several model variants tailored for different applications and computational requirements, NeuroSTORM-LowRes for lower spatial resolution inputs, NeuroSTORM-LongSeq for long sequences, NeuroSTORM-Base as the default configuration, NeuroSTORM-Large with an increased model capacity. In the supplementary Tab. 6, we detail the input size, hidden size, depths, and the corresponding parameter and computation costs for these variants.

\textbf{Spatiotemporal Redundancy Dropout module}. The STRD module is designed to address the inherent redundancy in fMRI data, which have slow spatial and temporal updates and contains significant context redundancy. By selectively masking redundant spatiotemporal information in masked image modeling paradigm ~\cite{xie2023data,he2022masked}, STRD encourages the model to focus on capturing complex long-range relationships within 4D fMRI sequences. In our approach, both spatial and temporal neighborhood redundancies are randomly dropped. For spatial contexts, if the model can reconstruct using temporal information, spatial redundancy is reduced through attention dropout. Conversely, for temporal contexts, if spatial information suffices for reconstruction, temporal redundancy is dropped. This dual approach ensures the model learns to model long-range spatiotemporal dependencies in fMRI data rather than relying on local redundancies. Mathematically, assume the input fMRI volume has dimensions $w \times h \times l \times t$. After patchification, it becomes $w' \times h' \times l' \times t'$, yielding $N_p$ patches. The attention matrix $A$ therefore lies in $\mathbb{R}^{N_p \times N_p}$. We define the spatial matching probability $f_{\text{spat}}(i)$ and the temporal matching probability $f_{\text{temp}}(i)$ as follows:

\begin{equation}
f_{\text{spat}}(i) = \max_{j \in \Omega_s(i)} (\hat{A}_{i,j}), \quad f_{\text{temp}}(i) = \max_{j \in \Omega_t(i)} (\hat{A}_{i,j})
\end{equation}

\noindent where \(\hat{A} = \text{softmax}_{\text{row}}(A)\), and \(\Omega_s(i)\) and \(\Omega_t(i)\) are the spatial and temporal index sets, respectively, which are a cubic window of edge length $l$ centred on the query voxel $i$:

\begin{equation}
    \begin{aligned}
        \Omega_s(i) &= \left\{\,j \;\middle|\; \max\bigl(|x_j-x_i|,\;|y_j-y_i|,\;|z_j-z_i|\bigr)
        \le \left\lfloor\frac{l}{2}\right\rfloor,\; j\neq i\right\}; \\
        \Omega_t(i) &= \left\{\,j \;\middle|\; |t_j-t_i|
        \le \left\lfloor\frac{l}{2}\right\rfloor,\; j\neq i\right\}.
    \end{aligned}
\end{equation}

The dropout probability \( W_{i,j} \) for each attention element is computed as:

\begin{equation}
W_{i,j} = \frac{1}{2} \left( \frac{f_{\text{temp}}(i) \hat{A}_{i,j}}{\sum_{j \in \Omega_s(i)} \hat{A}_{i,j}} + \frac{f_{\text{spat}}(i) \hat{A}_{i,j}}{\sum_{j \in \Omega_t(i)} \hat{A}_{i,j}} \right)
\end{equation}

By applying STRD, we prevent the model from exploiting redundant information for reconstruction, pushing it to learn meaningful spatiotemporal representations essential for understanding fMRI data.

\textbf{Task-specific Prompt Tuning}. In our TPT approach, we introduce learnable prompt parameters tailored for each downstream task while keeping the backbone parameters fixed. This strategy involves integrating a small set of task-specific prompts into the model's input space, allowing the network to adapt to new tasks with minimal data requirements.

For a given task, let the prompts be represented by a matrix \( P \in \mathbb{R}^{k \times d} \), where \( k \) is the number of prompts and \( d \) is the dimensionality of each prompt vector. These prompts are inserted at the input layer of the Transformer, influencing the model's processing without altering the core architecture.

The modified input to the first Transformer layer \( L_1 \) can be expressed as:

\begin{equation}
[x_1, Z_1] = L_1([x_0, P, E_0])
\end{equation}

\noindent where \( x_0 \) is the original input embedding, \( E_0 \) is the positional encoding, and \( Z_1 \) is the output of the first layer. Subsequent layers process the output as follows:

\begin{equation}
[x_{i}, Z_{i}] = L_{i}(x_{i-1}, Z_{i-1}, E_{i-1}), \quad i = 2, 3, \ldots, N 
\end{equation}

This formulation allows the model to leverage pre-trained knowledge while adapting to specific tasks through the learnable prompts \( P \), which are the only parameters updated during training. The backbone remains unchanged, ensuring computational efficiency and preserving the model's transferability across different tasks.

\textbf{Computational Performance}
NeuroSTORM demonstrated efficient resource utilization throughout its training pipeline. The pre-training phase was conducted on 4*A6000 GPUs with 48GB memory each, using a batch size of 4*8 over 30 epochs. This pre-training process required approximately 13 days to complete, with GPU memory consumption reaching 44.34GB per device. The substantial memory requirements during pre-training primarily stem from decoder training. For fine-tuning on downstream tasks, where classification/regression heads require relatively minimal computation, the processing time scaled linearly with dataset size. When fine-tuning on the HCP-YA dataset (batch size 4*8 for 20 epochs), the model completed in 5.36 hours while maintaining manageable memory usage at 7.46GB per GPU. Notably, our implementation leveraged SSD storage to eliminate I/O bottlenecks when processing full fMRI volumes during training.

Comparative analysis with SwiFT, another volume-based approach, revealed NeuroSTORM's practical advantages. In HCP-YA fine-tuning experiments, NeuroSTORM required significantly less GPU memory (7.46GB vs. 19.01GB) despite a modest 14.6\% speed reduction compared to SwiFT. This trade-off originates from Mamba's selective state mechanism, which processes sequences recursively rather than through full parallelization. While slightly impacting training speed, the memory efficiency enables NeuroSTORM to process longer temporal sequences in single passes. It is a crucial capability for task fMRI analysis where extended context capture is essential.

\textbf{Study Population}. We utilized a total of 12 diverse datasets, with 5 (UKB ~\cite{palmer2007uk}, ABCD ~\cite{casey2018adolescent}, HCP-YA~\cite{van2013wu}, HCP-A~\cite{van2013wu}, HCP-D~\cite{van2013wu}) used for pre-training and age/gender prediction task, 7 (HCP-EP~\cite{van2013wu}, ADHD200 ~\cite{brown2012adhd}, UCLA~\cite{poldrack2016phenome}, COBRE~\cite{calhoun2012exploring}, MND~\cite{chang2025fmri}, TCP ~\cite{chopra2024transdiagnostic}, DMT-HAR-MED~\cite{meling2024meditating}),  only for the fMRI analysis benchmark. During pre-training, we exclusively utilized unlabeled fMRI data, without incorporating any auxiliary labels or phenotypic information. For downstream tasks, we split each dataset into training, validation, and test sets, typically following an 8:1:1 ratio unless otherwise specified. Model selection was based on performance on the validation set, and the checkpoint with the lowest validation loss (or highest relevant metric) was subsequently evaluated on the test set. To ensure reproducibility and facilitate future research, we have provided standardized data split files for each dataset within our public code repository. In the supplementary Fig. 1, we provide visualizations of the dataset and phenotype label distributions to facilitate understanding of the dataset scale and the challenge of phenotype prediction task.

The UK Biobank (UKB)~\cite{palmer2007uk} is a comprehensive study from the UK, examining genetic and non-genetic influences on diseases in a cohort of about 60,000 middle-aged individuals. In our study, we utilized data from 40,842 of these participants. Its rich dataset includes fMRI data with a spatial resolution of $2.4 \text{mm} \times 2.4 \text{mm} \times 2.4 \text{mm}$ and a repetition time (TR) of $735ms$, offering invaluable insights for brain research. The Adolescent Brain Cognitive Development (ABCD) ~\cite{casey2018adolescent} Study, funded by the NIH, tracks brain development in 9,448 children across the US, providing multimodal data from neuroimaging to behavioral assessments at a resolution of $2.4 \text{mm} \times 2.4 \text{mm} \times 2.4 \text{mm}$ and $\text{TR}=800ms$. The Human Connectome Project (HCP)~\cite{van2013wu,glasser2013minimal} datasets used in pre-training include HCP-YA (1,206 subjects, $\text{TR}=720ms$)), HCP-D (652 subjects, $\text{TR}=800ms$)), and HCP-A (725 subjects, $\text{TR}=800ms$)), mapping brain connectivity across different life stages (young adults to aging populations) with high-resolution ($2 \text{mm} \times 2 \text{mm} \times 2 \text{mm}$ imaging data. The HCP-EP dataset focuses on early psychosis, capturing data from 252 subjects with the spatial resolution of $2 \text{mm} \times 2 \text{mm} \times 2 \text{mm}$ and $\text{TR}=800ms$. ADHD200~\cite{brown2012adhd} provides insights into ADHD with fMRI data at $3 \text{mm} \times 3 \text{mm} \times 4 \text{mm}$ and $\text{TR}=2000ms$ from 973 children and adolescents. UCLA~\cite{poldrack2016phenome} dataset includes neuroimaging at $3 \text{mm} \times 3 \text{mm} \times 4 \text{mm}$ and $\text{TR}=2000ms$ from 272 individuals with various psychiatric conditions. The COBRE~\cite{calhoun2012exploring} dataset offers 173 imaging data at $3.75 \text{mm} \times 3.75 \text{mm} \times 4.55 \text{mm}$ and $\text{TR}=2000ms$ on schizophrenia, and the TCP~\cite{chopra2024transdiagnostic} dataset spans multiple psychiatric diagnoses with 59 subjects at $2 \text{mm} \times 2 \text{mm} \times 2 \text{mm}$ and and $\text{TR}=800ms$. The MND~\cite{chang2025fmri} dataset explores neurological decline in patients diagnosed with Amyotrophic Lateral Sclerosis (ALS), with imaging from 59 participants (44 males, 15 females, including 36 ALS patients and 23 controls) collected at Herston Imaging Research Facility, Australia, using $2.395 \text{mm} \times 2.395 \text{mm} \times 2.4 \text{mm}$ and $\text{TR}=2000ms$ protocols on a 3T Siemens Prisma scanner. Collectively, these datasets facilitate research into brain function and psychiatric conditions, offering a robust foundation for developing advanced analytical models.

\textbf{fMRI Preprocessing}. For all datasets, the initial steps involve standardizing data to the Montreal Neurological Institute (MNI) space and spatially unifying each 3D volume to a consistent spatial dimension of $96 \times 96 \times 96$ with a spatial resolution of $2 \text{mm} \times 2 \text{mm} \times 2 \text{mm}$ and a repetition time of 0.8 seconds. If the original spatial resolution or TR of an fMRI sequence does not match the target specifications, we first perform resampling to the target resolution. Subsequently, we apply cropping or padding operations to standardize all data to the shape of $96 \times 96 \times 96$. For ROI-based methods, functional connectivities are generated using four atlases, AAL3, CC200, Harvard-Oxford, and Desikan-Killiany. The atlas originally employed by each comparative method is preferred. The resulting correlation matrices undergo Fisher transformation to improve normality, completing a standardized preprocessing pipeline that ensures both geometric compatibility and statistical comparability across diverse datasets. This foundational preprocessing facilitates robust comparison and analysis across studies.

The preprocessing of these datasets involves several standardized steps to ensure data quality and consistency across studies ~\cite{esteban2019fmriprep}. For instance, the ADHD200 and HCP-EP datasets undergo preprocessing that includes motion correction, normalization to the MNI space, and artifact removal, facilitating reliable analyses of brain connectivity. Researchers frequently utilize atlases such as the Schaefer atlas for extracting ROI-timeseries matrices, which are instrumental in studying functional connectivity and network integrity. The datasets support tasks like (reported) gender classification and age regression, leveraging their comprehensive phenotypic and demographic data. Advanced models, including BrainGNN ~\cite{li2021braingnn}, have been trained on these datasets to predict clinical outcomes and understand cognitive processes. The TCP dataset is processed using the HCP pipelines, including ICA-FIX denoising and global signal regression to control for noise. It provides analysis-ready functional connectivity matrices and supports the exploration of brain-behavior relationships across traditional diagnostic boundaries.

For state classification, we adopt the experimental setup from ~\cite{nguyen2020attend} to ensure a fair comparison. This involves ensuring that each data instance contains the same number of timesteps by extracting sets of \( k \) contiguous frames and looping the time series if fewer than \( k \) frames are present. During training, random frame sets are utilized, while validation and testing are conducted using the first \( k \) frames, where \( k = 40 \) in our experiments. We discuss the impact of different values of \( k \) on the experimental results in the appendix. Additionally, our NeuroSTORM framework is capable of directly processing complete tfMRI sequences.

\textbf{Label Preprocessing}. In our study, to provide appropriate supervision signals for models across various downstream tasks, label preprocessing is customized according to specific requirements. For classification tasks, including gender classification, disease diagnosis, and task fMRI state classification, we utilize one-hot encoding for the labels. This encoding converts categorical labels into a binary matrix representation, facilitating efficient model training and enhancing classification performance. For regression tasks, such as age prediction and phenotype prediction, we apply normalization ~\cite{kim2023swift,malkiel2022self} to all target variables. This process involves scaling the data to a standard range, improving model convergence and stability. During inference, we can rescale the predicted values back to their original magnitude using the recorded normalization parameters. This multifaceted approach to label preprocessing ensures that our models are well-equipped to handle the diverse challenges posed by both classification and regression tasks in neuroimaging analysis.

\textbf{Implementation Details.} All foundation models, including NeuroSTORM, SwiFT~\cite{kim2023swift}, TFF~\cite{malkiel2022self}, BrainLM~\cite{ortega2023brainlm}, and Brain-JEPA ~\cite{dong2024brain}, are first trained with self-supervised pretraining and then fine-tuned on each downstream task. For these models we adopt the official code bases and released pretrained weights for network definition and initialization. The remaining baselines~\cite{li2021braingnn,kan2022brain,zhang2022classification,bannadabhavi2023community,cui2022interpretable} are trained from scratch with fully supervised learning. All ROI-based method use the atlas specified in their original paper. All compared methods follow an identical 8:1:1 split of the data into training, validation and test sets. We also unify the remaining hyperparameters: on 4$\times$A6000 GPUs, all models are trained with batch size $48$. Training uses the DDP strategy, learning rate $5\times10^{-5}$, gradient clipping, the AdamW optimizer with weight decay $0.01$, and a CosineAnnealingWarmUp schedule with $\gamma=1.0$ and cycle ratio $0.3$.

\section{Data availability}
All data used in this study were obtained from publicly available or restricted-access datasets: UK Biobank (\url{https://www.ukbiobank.ac.uk/}); ABCD study (\url{https://nda.nih.gov/abcd}); Human Connectome Project (\url{https://db.humanconnectome.org/}); ADHD200 (\url{https://fcon_1000.projects.nitrc.org/indi/adhd200/}); ABIDE (\url{https://fcon_1000.projects.nitrc.org/indi/abide/}); UCLA (\url{https://openneuro.org/datasets/ds000030}); COBRE (\url{https://fcon_1000.projects.nitrc.org/indi/retro/cobre.html}); HBN (\url{https://fcon_1000.projects.nitrc.org/indi/cmi_healthy_brain_network/}); PNC ( \url{https://www.med.upenn.edu/bbl/philadelphianeurodevelopmentalcohort.html}, with controlled access typically via dbGaP \url{https://www.ncbi.nlm.nih.gov/gap/}); REST-meta-MDD (\url{http://rfmri.org/REST-meta-MDD}); DMT-HAR-MED (\url{https://openneuro.org/datasets/ds006644/versions/1.0.1}; TCP (\url{https://openneuro.org/datasets/ds005237/versions/1.1.3}); and MND (\url{https://openneuro.org/datasets/ds005874/versions/1.1.0}).

\section{Code availability}
The NeuroSTORM project homepage is available at \url{https://cuhk-aim-group.github.io/NeuroSTORM/}. All code used in this study, including the NeuroSTORM implementation as well as scripts for preprocessing, pre-training, fine-tuning, and benchmarking across downstream tasks, is publicly available at \url{https://github.com/CUHK-AIM-Group/NeuroSTORM}.

\section{Acknowledgments}
This work was supported by NSFC/RGC Joint Research Scheme N\_CUHK4126/25, Innovation and Technology Fund Mainland-Hong Kong Joint Funding Scheme MHP/173/24 and Major Program of National Natural Science Foundation of China (No.62394314)

\section{Author contributions}
C.W. contributed to the development of methods, analysis and interpretation of the data, and drafting of the manuscript. Y.J. and Z.P. contributed to figure preparation and drafting of the manuscript. C.L. contributed to the development of methods and the experimental design. C.B., L.Z., J.L., L.H. and B.-H.K. contributed to fMRI data collection and preprocessing. W.F., J.S., C.Y., T.L., X.-J.K., Q.L., D.B., A.Q. and R.H. provided analytic input. H.H., X.L. and Y.Y. provided funding, oversaw the analysis and interpretation of the data, and contributed to the drafting of the manuscript.

\section{Competing interests}
The authors declare no competing interests.

\clearpage

\section{References}
\begin{spacing}{0.9}
\bibliographystyle{naturemag}


\begin{thebibliography}{99}

\bibitem{hearne2016functional}
Luke~J. Hearne, Jason~B. Mattingley, and Luca Cocchi.
\newblock Functional Brain Networks Related to Individual Differences in Human Intelligence at Rest.
\newblock \emph{Scientific Reports}, 6(1):1--8, 2016.

\bibitem{sorg2007selective}
Christian Sorg, Valentin Riedl, Mark M{\"u}hlau, Vince~D. Calhoun, Tom Eichele,
Leonhard L{\"a}er, Alexander Drzezga, Hans F{\"o}rstl, Alexander Kurz, Claus Zimmer,
et~al.
\newblock Selective Changes of Resting-State Networks in Individuals at Risk for Alzheimer's Disease.
\newblock \emph{Proceedings of the National Academy of Sciences}, 104(47):18760--18765, 2007.

\bibitem{lewandowski2010polyamine}
Nicole~M. Lewandowski, Shulin Ju, Miguel Verbitsky, Barbara Ross, Melissa~L. Geddie,
Edward Rockenstein, Anthony Adame, Alim Muhammad, Jean~Paul Vonsattel, Dagmar Ringe,
et~al.
\newblock Polyamine Pathway Contributes to the Pathogenesis of Parkinson Disease.
\newblock \emph{Proceedings of the National Academy of Sciences}, 107(39):16970--16975, 2010.

\bibitem{scotti2024reconstructing}
Paul Scotti, Atmadeep Banerjee, Jimmie Goode, Stepan Shabalin, Alex Nguyen, Aidan Dempster,
Nathalie Verlinde, Elad Yundler, David Weisberg, Kenneth Norman, et~al.
\newblock Reconstructing the mind's eye: fMRI-to-image with contrastive learning and diffusion priors.
\newblock \emph{Advances in Neural Information Processing Systems}, 36, 2024.

\bibitem{he2022meta}
Tong He, Lijun An, Pansheng Chen, Jianzhong Chen, Jiashi Feng, Danilo Bzdok,
Avram~J. Holmes, Simon~B. Eickhoff, and B.~T. Thomas Yeo.
\newblock Meta-matching as a simple framework to translate phenotypic predictive models from big to small data.
\newblock \emph{Nature Neuroscience}, 25(6):795--804, 2022.

\bibitem{kim2023large}
Byung-Hoon Kim, Jungwon Choi, EungGu Yun, Kyungsang Kim, Xiang Li, and Juho Lee.
\newblock Large-scale graph representation learning of dynamic brain connectome with transformers.
\newblock \emph{arXiv preprint} arXiv:2312.14939, 2023.

\bibitem{kim2024learning}
Byung-Hoon Kim, Jungwon Choi, EungGu Yun, Kyungsang Kim, Xiang Li, and Juho Lee.
\newblock Learning dynamic brain connectome with graph transformers for psychiatric diagnosis classification.
\newblock In \emph{2024 IEEE International Symposium on Biomedical Imaging (ISBI)}, pages 1--5. IEEE, 2024.

\bibitem{ortega2023brainlm}
Josue Ortega Caro, Antonio~H. Oliveira Fonseca, Christopher Averill, Syed~A. Rizvi,
Matteo Rosati, James~L. Cross, Prateek Mittal, Emanuele Zappala, Daniel Levine,
and Rahul~M. Dhodapkar.
\newblock BrainLM: A foundation model for brain activity recordings.
\newblock \emph{bioRxiv}, 2023. doi:10.1101/2023.09.12.557460.

\bibitem{zhang2021identification}
Yu Zhang, Wei Wu, Russell~T. Toll, Sharon Naparstek, Adi Maron-Katz, Mallissa Watts,
Joseph Gordon, Jisoo Jeong, Laura Astolfi, Emmanuel Shpigel, et~al.
\newblock Identification of psychiatric disorder subtypes from functional connectivity patterns in resting-state electroencephalography.
\newblock \emph{Nature Biomedical Engineering}, 5(4):309--323, 2021.

\bibitem{zhang2022predicting}
Lu Zhang, Li Wang, Dajiang Zhu, Alzheimer's Disease Neuroimaging Initiative, et~al.
\newblock Predicting brain structural network using functional connectivity.
\newblock \emph{Medical Image Analysis}, 79:102463, 2022. doi:10.1016/j.media.2022.102463.

\bibitem{zhang2024self}
Kexin Zhang, Qingsong Wen, Chaoli Zhang, Rongyao Cai, Ming Jin, Yong Liu, James~Y. Zhang,
Yuxuan Liang, Guansong Pang, Dongjin Song, and Shirui Pan.
\newblock Self-supervised learning for time series analysis: Taxonomy, progress, and prospects.
\newblock \emph{IEEE Transactions on Pattern Analysis and Machine Intelligence}, 46(10):6775--6794, 2024.
\newblock doi:10.1109/TPAMI.2024.3387317.

\bibitem{tian2020topographic}
Ye Tian, Daniel~S. Margulies, Michael Breakspear, and Andrew Zalesky.
\newblock Topographic Organization of the Human Subcortex Unveiled with Functional Connectivity Gradients.
\newblock \emph{Nature Neuroscience}, 23(11):1421--1432, 2020.

\bibitem{biswal2025history}
Bharat~B. Biswal and Lucina~Q. Uddin.
\newblock The history and future of resting-state functional magnetic resonance imaging.
\newblock \emph{Nature}, 641(8065):1121--1131, 2025.

\bibitem{botviniknezer2023reproducibility}
Rotem Botvinik-Nezer and Tor~D. Wager.
\newblock Reproducibility in neuroimaging analysis: Challenges and solutions.
\newblock \emph{Biological Psychiatry: Cognitive Neuroscience and Neuroimaging}, 8(8):780--788, 2023.
\newblock doi:10.1016/j.bpsc.2022.12.006.

\bibitem{cao2014test}
Hengyi Cao, Michael~M. Plichta, Axel Sch{\"a}fer, Leila Haddad, Oliver Grimm, Michael Schneider,
Christine Esslinger, Peter Kirsch, Andreas Meyer-Lindenberg, and Heike Tost.
\newblock Test--retest reliability of fMRI-based graph theoretical properties during working memory, emotion processing, and resting state.
\newblock \emph{NeuroImage}, 84:888--900, 2014. doi:10.1016/j.neuroimage.2013.09.013.

\bibitem{waller2022enigma}
Lea Waller, Susanne Erk, Elena Pozzi, Yara~J. Toenders, Courtney~C. Haswell, Marc B{\"u}ttner,
Paul~M. Thompson, Lianne Schmaal, Rajendra~A. Morey, Henrik Walter, et~al.
\newblock ENIGMA HALFpipe: Interactive, Reproducible, and Efficient Analysis for Resting-State and Task-Based fMRI Data.
\newblock \emph{Human Brain Mapping}, 43(9):2727--2742, 2022.

\bibitem{noble2021guide}
Stephanie Noble, Dustin Scheinost, and Robert~Todd Constable.
\newblock A Guide to the Measurement and Interpretation of fMRI Test-Retest Reliability.
\newblock \emph{Current Opinion in Behavioral Sciences}, 40:27--32, 2021.

\bibitem{achiam2023gpt}
Josh Achiam, Steven Adler, Sandhini Agarwal, Lama Ahmad, Ilge Akkaya, Florencia~Leoni Aleman,
Diogo Almeida, Janko Altenschmidt, Sam Altman, Shyamal Anadkat, et~al.
\newblock GPT-4 Technical Report.
\newblock \emph{arXiv preprint} arXiv:2303.08774, 2023.

\bibitem{touvron2023llama}
Hugo Touvron, Thibaut Lavril, Gautier Izacard, Xavier Martinet, Marie-Anne Lachaux,
Timoth{\'e}e Lacroix, Baptiste Rozi{\`e}re, Naman Goyal, Eric Hambro, Faisal Azhar, et~al.
\newblock Llama: Open and Efficient Foundation Language Models.
\newblock \emph{arXiv preprint} arXiv:2302.13971, 2023.

\bibitem{xu2024whole}
Hanwen Xu, Naoto Usuyama, Jaspreet Bagga, Sheng Zhang, Rajesh Rao, Tristan Naumann, Cliff Wong,
Zelalem Gero, Javier Gonz{\'a}lez, Yu Gu, et~al.
\newblock A Whole-Slide Foundation Model for Digital Pathology from Real-World Data.
\newblock \emph{Nature}, pages 1--8, 2024.

\bibitem{zhou2023foundation}
Yukun Zhou, Mark~A. Chia, Siegfried~K. Wagner, Murat~S. Ayhan, Dominic~J. Williamson, Robbert~R. Struyven,
Timing Liu, Moucheng Xu, Mateo~G. Lozano, Peter Woodward-Court, et~al.
\newblock A Foundation Model for Generalizable Disease Detection from Retinal Images.
\newblock \emph{Nature}, 622(7981):156--163, 2023. doi:10.1038/s41586-023-05306-5.

\bibitem{zhang2024challenges}
Shaoting Zhang and Dimitris Metaxas.
\newblock On the Challenges and Perspectives of Foundation Models for Medical Image Analysis.
\newblock \emph{Medical Image Analysis}, 91:102996, 2024. doi:10.1016/j.media.2023.102996.

\bibitem{dey2024anystar}
Neel Dey, Mazdak Abulnaga, Benjamin Billot, Esra Abaci Turk, Ellen Grant, Adrian~V. Dalca, and Polina Golland.
\newblock AnyStar: Domain Randomized Universal Star-Convex 3D Instance Segmentation.
\newblock In \emph{Proceedings of the IEEE/CVF Winter Conference on Applications of Computer Vision (WACV)},
pages 7593--7603, 2024. doi:10.1109/WACV57701.2024.00742.

\bibitem{kim2023swift}
Peter Kim, Junbeom Kwon, Sunghwan Joo, Sangyoon Bae, Donggyu Lee, Yoonho Jung, Shinjae Yoo, Jiook Cha, and Taesup Moon.
\newblock Swift: Swin 4d fmri transformer.
\newblock \emph{Advances in Neural Information Processing Systems}, 36:42015--42037, 2023.

\bibitem{malkiel2022self}
Itzik Malkiel, Gony Rosenman, Lior Wolf, and Talma Hendler.
\newblock Self-Supervised Transformers for fMRI Representation.
\newblock In \emph{Proceedings of the International Conference on Medical Imaging with Deep Learning (MIDL)},
pages 895--913, 2022. PMLR.

\bibitem{palmer2007uk}
Lyle~J. Palmer.
\newblock UK Biobank: bank on it.
\newblock \emph{The Lancet}, 369(9578):1980--1982, 2007.

\bibitem{casey2018adolescent}
Betty~Jo Casey, Tariq Cannonier, May~I. Conley, Alexandra~O. Cohen, Deanna~M. Barch,
Mary~M. Heitzeg, Mary~E. Soules, Theresa Teslovich, Danielle~V. Dellarco, Hugh Garavan, et~al.
\newblock The adolescent brain cognitive development (ABCD) study: imaging acquisition across 21 sites.
\newblock \emph{Developmental Cognitive Neuroscience}, 32:43--54, 2018.

\bibitem{van2013wu}
David~C. Van Essen, Stephen~M. Smith, Deanna~M. Barch, Timothy~E.~J. Behrens, Essa Yacoub,
Kamil Ugurbil, Wu-Minn HCP Consortium, et~al.
\newblock The WU-Minn Human Connectome Project: an overview.
\newblock \emph{NeuroImage}, 80:62--79, 2013. doi:10.1016/j.neuroimage.2013.05.041.

\bibitem{chopra2024transdiagnostic}
Sidhant Chopra, Carrisa~V. Cocuzza, Connor Lawhead, Jocelyn~A. Ricard, Lo{\"i}c Labache,
Lauren~M. Patrick, Poornima Kumar, Arielle Rubenstein, Julia Moses, Lia Chen, et~al.
\newblock The Transdiagnostic Connectome Project: a richly phenotyped open dataset for advancing the study of brain-behavior relationships in psychiatry.
\newblock \emph{medRxiv}, 2024. doi:10.1101/2024.06.18.24309054.

\bibitem{chang2025fmri}
Jeryn Chang, JingLei Lv, Christine~C. Guo, Diana Lucia, Saskia Bollmann, Kelly Garner, Pamela~A. McCombe,
Robert~D. Henderson, Thomas~B. Shaw, Frederik~J. Steyn, et~al.
\newblock An fMRI Dataset for Appetite Neural Correlates in People Living with Motor Neuron Disease.
\newblock \emph{Scientific Data}, 12(1):466, 2025.

\bibitem{brown2012adhd}
Matthew~R.~G. Brown, Gagan~S. Sidhu, Russell Greiner, Nasimeh Asgarian, Meysam Bastani,
Peter~H. Silverstone, Andrew~J. Greenshaw, and Serdar~M. Dursun.
\newblock ADHD-200 Global Competition: diagnosing ADHD using personal characteristic data can outperform resting state fMRI measurements.
\newblock \emph{Frontiers in Systems Neuroscience}, 6:69, 2012. doi:10.3389/fnsys.2012.00069.

\bibitem{calhoun2012exploring}
Vince~D. Calhoun, Jing Sui, Kent Kiehl, Jessica Turner, Elena Allen, and Godfrey Pearlson.
\newblock Exploring the Psychosis Functional Connectome: Aberrant Intrinsic Networks in Schizophrenia and Bipolar Disorder.
\newblock \emph{Frontiers in Psychiatry}, 2:75, 2012. doi:10.3389/fpsyt.2011.00075.

\bibitem{poldrack2016phenome}
Russell~A. Poldrack, Eliza Congdon, William Triplett, Krzysztof~J. Gorgolewski, Katherine~H. Karlsgodt,
Jonathan~A. Mumford, Farrah~W. Sabb, Nelson~B. Freimer, Edythe~D. London, Tyrone~D. Cannon, et~al.
\newblock A Phenome-wide Examination of Neural and Cognitive Function.
\newblock \emph{Scientific Data}, 3:160110, 2016. doi:10.1038/sdata.2016.110.

\bibitem{meling2024meditating}
Daniel Meling, Klemens Egger, Helena~D. Aicher, Javier Jare{\~n}o Redondo, Jovin Mueller,
Jo{\"e}lle Dornbierer, Elijah Temperli, Emilia~A. Vasella, Luzia Caflisch, David~J. Pfeiffer, et~al.
\newblock Meditating on psychedelics. A randomized placebo-controlled study of DMT and harmine in a mindfulness retreat.
\newblock \emph{Journal of Psychopharmacology}, 38(10):897--910, 2024.

\bibitem{li2021braingnn}
Xiaoxiao Li, Yuan Zhou, Nicha Dvornek, Muhan Zhang, Siyuan Gao, Juntang Zhuang, Dustin Scheinost,
Lawrence~H. Staib, Pamela Ventola, and James~S. Duncan.
\newblock Braingnn: Interpretable brain graph neural network for fmri analysis.
\newblock \emph{Medical Image Analysis}, 74:102233, 2021. doi:10.1016/j.media.2021.102233.

\bibitem{kan2022brain}
Xuan Kan, Wei Dai, Hejie Cui, Zilong Zhang, Ying Guo, and Carl Yang.
\newblock Brain Network Transformer.
\newblock \emph{Advances in Neural Information Processing Systems}, 35:25586--25599, 2022.

\bibitem{zhang2022classification}
Hao Zhang, Ran Song, Liping Wang, Lin Zhang, Dawei Wang, Cong Wang, and Wei Zhang.
\newblock Classification of brain disorders in rs-fMRI via local-to-global graph neural networks.
\newblock \emph{IEEE Transactions on Medical Imaging}, 42(2):444--455, 2023. doi:10.1109/TMI.2022.3219260.

\bibitem{bannadabhavi2023community}
Anushree Bannadabhavi, Soojin Lee, Wenlong Deng, Rex Ying, and Xiaoxiao Li.
\newblock Community-Aware Transformer for Autism Prediction in fMRI Connectome.
\newblock In \emph{Proceedings of the International Conference on Medical Image Computing and Computer-Assisted Intervention (MICCAI)},
pages 287--297. Springer, 2023.

\bibitem{cui2022interpretable}
Hejie Cui, Wei Dai, Yanqiao Zhu, Xiaoxiao Li, Lifang He, and Carl Yang.
\newblock Interpretable graph neural networks for connectome-based brain disorder analysis.
\newblock In \emph{Proceedings of the International Conference on Medical Image Computing and Computer-Assisted Intervention (MICCAI)},
pages 375--385. Springer, 2022.

\bibitem{dong2024brain}
Zijian Dong, Ruilin Li, Yilei Wu, Thuan~Tinh Nguyen, Joanna Chong, Fang Ji, Nathanael Tong,
Christopher Chen, and Juan~Helen Zhou.
\newblock Brain-jepa: Brain dynamics foundation model with gradient positioning and spatiotemporal masking.
\newblock \emph{Advances in Neural Information Processing Systems}, 37:86048--86073, 2024.

\bibitem{jablensky2016psychiatric}
Assen Jablensky.
\newblock Psychiatric classifications: validity and utility.
\newblock \emph{World Psychiatry}, 15(1):26--31, 2016.

\bibitem{joyce2023explainable}
Dan~W. Joyce, Andrey Kormilitzin, Katharine~A. Smith, and Andrea Cipriani.
\newblock Explainable artificial intelligence for mental health through transparency and interpretability for understandability.
\newblock \emph{npj Digital Medicine}, 6(1):6, 2023.

\bibitem{finn2015functional}
Emily~S. Finn, Xilin Shen, Dustin Scheinost, Monica~D. Rosenberg, Jessica Huang, Marvin~M. Chun,
Xenophon Papademetris, and R.~Todd Constable.
\newblock Functional connectome fingerprinting: identifying individuals using patterns of brain connectivity.
\newblock \emph{Nature Neuroscience}, 18(11):1664--1671, 2015.

\bibitem{nguyen2020attend}
Sam Nguyen, Brenda Ng, Alan~D. Kaplan, and Priyadip Ray.
\newblock Attend and Decode: 4D fMRI Task State Decoding Using Attention Models.
\newblock In \emph{Machine Learning for Health}, pages 267--279. PMLR, 2020.

\bibitem{alexander2017open}
Lindsay~M. Alexander, Jasmine Escalera, Lei Ai, Charissa Andreotti, Karina Febre, Alexander Mangone,
Natan Vega-Potler, Nicolas Langer, Alexis Alexander, Meagan Kovacs, et~al.
\newblock An open resource for transdiagnostic research in pediatric mental health and learning disorders.
\newblock \emph{Scientific Data}, 4(1):1--26, 2017.

\bibitem{satterthwaite2014neuroimaging}
Theodore~D. Satterthwaite, Mark~A. Elliott, Kosha Ruparel, James Loughead, Karthik Prabhakaran,
Monica~E. Calkins, Ryan Hopson, Chad Jackson, Jack Keefe, Marisa Riley, et~al.
\newblock Neuroimaging of the Philadelphia neurodevelopmental cohort.
\newblock \emph{NeuroImage}, 86:544--553, 2014.

\bibitem{yan2019reduced}
Chao-Gan Yan, Xiao Chen, Le Li, Francisco~Xavier Castellanos, Tong-Jian Bai, Qi-Jing Bo, Jun Cao,
Guan-Mao Chen, Ning-Xuan Chen, Wei Chen, et~al.
\newblock Reduced default mode network functional connectivity in patients with recurrent major depressive disorder.
\newblock \emph{Proceedings of the National Academy of Sciences}, 116(18):9078--9083, 2019.

\bibitem{cao2022swin}
Hu Cao, Yueyue Wang, Joy Chen, Dongsheng Jiang, Xiaopeng Zhang, Qi Tian, and Manning Wang.
\newblock Swin-Unet: Unet-like Pure Transformer for Medical Image Segmentation.
\newblock In \emph{Proceedings of the European Conference on Computer Vision (ECCV) Workshops}, pages 205--218, 2022.
\newblock doi:10.1007/978-3-031-25066-8\_9.

\bibitem{gu2021efficiently}
Albert Gu, Karan Goel, and Christopher R{\'e}.
\newblock Efficiently Modeling Long Sequences with Structured State Spaces.
\newblock \emph{arXiv preprint} arXiv:2111.00396, 2021.

\bibitem{gu2021combining}
Albert Gu, Isys Johnson, Karan Goel, Khaled Saab, Tri Dao, Atri Rudra, and Christopher R{\'e}.
\newblock Combining Recurrent, Convolutional, and Continuous-Time Models with Linear State Space Layers.
\newblock \emph{Advances in Neural Information Processing Systems}, 34:572--585, 2021.

\bibitem{gu2023mamba}
Albert Gu and Tri Dao.
\newblock Mamba: Linear-time Sequence Modeling with Selective State Spaces.
\newblock \emph{arXiv preprint} arXiv:2312.00752, 2023.

\bibitem{xie2023data}
Zhenda Xie, Zheng Zhang, Yue Cao, Yutong Lin, Yixuan Wei, Qi Dai, and Han Hu.
\newblock On Data Scaling in Masked Image Modeling.
\newblock In \emph{Proceedings of the IEEE/CVF Conference on Computer Vision and Pattern Recognition (CVPR)},
pages 10365--10374, 2023.

\bibitem{he2022masked}
Kaiming He, Xinlei Chen, Saining Xie, Yanghao Li, Piotr Doll{\'a}r, and Ross Girshick.
\newblock Masked Autoencoders are Scalable Vision Learners.
\newblock In \emph{Proceedings of the IEEE/CVF Conference on Computer Vision and Pattern Recognition (CVPR)},
pages 16000--16009, 2022.

\bibitem{glasser2013minimal}
Matthew~F. Glasser, Stamatios~N. Sotiropoulos, J.~Anthony Wilson, Timothy~S. Coalson, Bruce Fischl,
Jesper~L. Andersson, Junqian Xu, Saad Jbabdi, Matthew Webster, Jonathan~R. Polimeni, et~al.
\newblock The minimal preprocessing pipelines for the Human Connectome Project.
\newblock \emph{NeuroImage}, 80:105--124, 2013.

\bibitem{esteban2019fmriprep}
Oscar Esteban, Christopher~J. Markiewicz, Ross~W. Blair, Craig~A. Moodie, A.~Ilkay Isik, Asier Erramuzpe,
James~D. Kent, Mathias Goncalves, Elizabeth DuPre, Madeleine Snyder, et~al.
\newblock fMRIPrep: a robust preprocessing pipeline for functional MRI.
\newblock \emph{Nature Methods}, 16(1):111--116, 2019. doi:10.1038/s41592-018-0235-4.

\end{thebibliography}
\end{spacing}

\end{document}